\documentclass[runningheads]{llncs}
\usepackage[T1]{fontenc}
\usepackage{graphicx,verbatim}
\graphicspath{ {./figures/} }
\usepackage{subcaption}
\usepackage{booktabs}
\usepackage{multirow}
\usepackage{appendix}
\usepackage{cite}
\usepackage{hyperref}
\usepackage{color}

\usepackage{amsmath}
\usepackage{amssymb}
\usepackage[normalem]{ulem}

\begin{document}
\title{Influence of Classification Task and Distribution Shift Type on OOD Detection in Fetal Ultrasound}
\titlerunning{Influence of Cls.~Task and Dist.~Shift on OOD Detection in Fetal Ultrasound}
%
\author{Chun Kit Wong\inst{1,2}\orcidID{0000-0001-5528-9727} \and
Anders N. Christensen\inst{1}\orcidID{0000-0002-3668-3128} \and
Cosmin I. Bercea\inst{3,4}\orcidID{0000-0003-2628-2766} \and
Julia A. Schnabel\inst{3,4,5}\orcidID{0000-0001-6107-3009} \and
Martin G. Tolsgaard\inst{6,7}\orcidID{0000-0001-9197-5564} \and
Aasa Feragen\inst{1,2}\orcidID{0000-0002-9945-981X}
}
%
\authorrunning{C. K. Wong et al.}
%
\institute{Technical University of Denmark, Kongens Lyngby, Denmark \\
    \email{\{ckwo,afhar\}@dtu.dk} \and
    Pioneer Centre for AI, Copenhagen, Denmark \and
    Technical University of Munich, Munich, Germany \and
    Helmholtz AI and Helmholtz Munich, Munich, Germany \and
    King's College London, London, UK \and
    University of Copenhagen, Copenhagen, Denmark \and
    CAMES Rigshospitalet, Copenhagen, Denmark 
}
    
\maketitle              
\begin{abstract}
Reliable out-of-distribution (OOD) detection is important for safe deployment of deep learning models in fetal ultrasound amidst heterogeneous image characteristics and clinical settings. OOD detection relies on estimating a classification model's uncertainty, which should increase for OOD samples. While existing research has largely focused on uncertainty quantification methods, this work investigates the impact of the classification task itself. Through experiments with eight uncertainty quantification methods across four classification tasks, we demonstrate that OOD detection performance significantly varies with the task, and that the best task depends on the defined ID-OOD criteria; specifically, whether the  OOD sample is due to: i) an image characteristic shift or ii) an anatomical feature shift. Furthermore, we reveal that superior OOD detection does not guarantee optimal abstained prediction, underscoring the necessity to align task selection and uncertainty strategies with the specific downstream application in medical image analysis.

\end{abstract}

\section{Introduction}
Out-of-distribution (OOD) detection is crucial for deploying reliable deep learning models in medical image analysis. This is particularly needed in fetal ultrasound, which is ubiquitous in routine maternity check-ups, but also comes with significant heterogeneity in image characteristics due to differences in operator training or maternal Body Mass Index (BMI), and a diverse range of ultrasound scanners. This variability in input image distribution directly impacts the performance of deep learning models, underscoring the need for robust OOD detection to identify distributional shifts of different kinds and ensure diagnostic accuracy.

\textbf{OOD detection finds wide application in medical imaging~\cite{hong2024out,zadorozhny2022out}}. For image quality and domain shift detection, \cite{koch2024distribution} attempted to detect distribution shift for surveillance of deployed AI algorithms, and \cite{li2022estimating} developed calibration technique to estimate performance of a trained model under domain shift. Another task is anatomical shift detection, e.g.~identifying unseen pathologies or abnormalities during inference. In chest X-ray analysis, \cite{berger2021confidence} identified fracture cases as OOD examples using a classifier trained to distinguish cardiomegaly from pneumothorax.  Similarly, in dermatology, \cite{combalia2020uncertainty} utilized images of unseen skin diseases as OOD sets. In digital pathology, researchers have explored the detection of novel abnormalities as OOD samples~\cite{thagaard2020can,linmans2020efficient}.  Beyond pathology detection, OOD methods are also used to extract clinically relevant frames from ultrasound videos, identifying frames that deviate from expected anatomical content~\cite{mishra2023dual}.

\textbf{OOD detection can be formulated via uncertainty quantification (UQ).} There is a diverse landscape of UQ techniques applicable to medical imaging~\cite{huang2024review,lambert2024trustworthy,zou2023review}, many focusing on image classification~\cite{kurz2022uncertainty,tardy2019uncertainty}. Here, OOD detection relies on estimating the predictive uncertainty of a classifier for a given input image, reflecting the model's confidence in its prediction. Higher uncertainty typically indicates a greater likelihood that the input image originates from a distribution different from the training data, suggesting it is out-of-distribution. 

While many easily-available UQ algorithms are based on how a particular classifier views the data, being OOD is a natural property of the data itself. This is important from the clinic's point-of-view, where in fetal ultrasound we observe both classical quality shifts such as blur and low resolution, or other shifts in image characteristics such as hue or artifacts -- but also anatomical shifts in off-plane images. In the clinic, it is therefore important that OOD detection is robust and consistent for different types of distribution shifts.

\textbf{Contributions.} We study eight common classifier-based UQ techniques for OOD detection. To study robustness of OOD detection across types of distribution shifts, we challenge the idea that the UQ techniques should be based on the \textbf{primary classifier of interest}, which for this paper will be an \emph{anatomical plane classifier}. As UQ base classifiers, we train three alternative models to predict image meta-information found in the DICOM header, and study how the resulting four models perform as a basis for the different UQ-methods, validated on OOD distribution for i) image characteristic distribution shifts, and ii) anatomical shifts. Finally, we test how our primary classifier of interest performs in an "abstained prediction" setting using the different OOD models.

We find that the choice of base classifier for the OOD detector has a large effect on OOD performance, and that the primary classifier is not always the best choice. However, we see that these results are surprisingly not indicative of performance in "abstained prediction" for the primary classifier -- where the UQ-methods based on the primary classifier perform better than the competitors. This shows that the choice of base classifiers matters for OOD detection, as well as that OOD detection and trustworthy classification are not always two sides of the same coin. Moreover, this leaves open questions about appropriate validation design for OOD detection.

\section{Method}
\subsection{Uncertainty Quantification (UQ) Methods}
\label{sec:uncertainty_quantification_methods}
Under a $C$ classes classification setting, a classifier is trained to predict the class label $\mathbf{y}\in\{1 \dots C\}$ given an input image $\mathbf{x}\in\mathbb{R}^2$, with a predictive uncertainty score $u(\mathbf{x})$ that can be estimated using UQ-methods. OOD detection is then achieved by thresholding: If $u(\mathbf{x}) > t$, then $\mathbf{x}$ is flagged as OOD. Inspired by~\cite{mucsanyi2024benchmarking}, we evaluate eight UQ-methods that do not rely on a hold-out OOD test set during training. We focus on the following, mainly deterministic, methods considering their fast, non-iterative inference procedure:

As a \textbf{Baseline}, we trained a ResNet-50 model for the classification task and calculated the entropy of the model's predictive softmax probability as $u(\mathbf{x})$. \textbf{Temperature Scaling}~\cite{guo2017calibration} adds a calibration step to the softmax probability.

Two alternative methods augment the baseline model with an auxiliary prediction head. \textbf{Loss Prediction}~\cite{yoo2019learning,lahlou2021deup,kirchhof2023url} trains this head to predict the loss $L(\mathbf{x})$, while \textbf{Correctness Prediction}~\cite{mucsanyi2024benchmarking,kirchhof2023url} trains it to predict the probability $p(\mathbf{\hat{y}} \neq \mathbf{y} | \mathbf{x})$ of an incorrect prediction, using the predicted values as $u(\mathbf{x})$.

Meanwhile, two methods make use of the feature embedding space density. With a trained classification model, \textbf{Deterministic Uncertainty Quantification (DUQ)}~\cite{van2020uncertainty} and \textbf{Deep Deterministic Uncertainty (DDU)}~\cite{mukhoti2023deep} work by first obtaining feature embeddings of all training images, followed by learning a density estimator using these embeddings. To ensure the latent space is well-regularized, DUQ adds a gradient penalty term in the loss function, while DDU applies spectral normalization to the model weights. $u(\mathbf{x})$ is then given by one minus the estimated density of $\mathbf{x}$ in the feature embedding space.

Finally, we also evaluated two probabilistic methods given their popularity in medical image analysis literature~\cite{lambert2024trustworthy}. In these methods, multiple predictions are obtained for a given $\mathbf{x}$. For each prediction, \textbf{MC-dropout}~\cite{gal2016dropout,srivastava2014dropout} switches off a random subset of the model activations, while \textbf{Ensemble}~\cite{lakshminarayanan2017simple} uses a trained model that is initialized differently. Here, we used a lightweight implementation of ensemble~\cite{lee2015m}, which involves training a single model with multiple randomly-initialized heads. Entropy of these predictions are taken as $u(\mathbf{x})$.

\subsection{Dataset}
\label{sec:dataset}
We utilized a combination of two public and one private fetal ultrasound datasets.

\textbf{The SONAI dataset} is a private fetal ultrasound dataset including images from four common fetal anatomical planes: abdomen, brain, femur, and thorax, acquired using advanced ultrasound scanners. Additionally, each image is accompanied by metadata in the form of a DICOM header, which we utilized in designing our classification tasks (see \autoref{sec:classification_tasks}). This dataset includes images of other `generic' fetal ultrasound planes, which we group into a separate bundle named \textbf{the SONAI (Other) dataset}. This will be used as an OOD dataset.

\textbf{The BCNatal dataset}\cite{burgos2020evaluation} also comprises images from the same four fetal anatomical plane, acquired using advanced ultrasound scanners similar to those used for the SONAI dataset. This dataset also includes images of other `generic' ultrasound images, where we group into a separate bundle and refer to as \textbf{the BCNatal (Other) dataset}, to be used as an OOD dataset.

\textbf{The African dataset}\cite{sendra2023generalisability} comprises fetal ultrasound images of the same four anatomical planes as the two other datasets, but acquired in resource-constrained settings with less advanced ultrasound scanners. This results in images of lower quality compared to the two other datasets.

\subsection{Classification tasks}
\label{sec:classification_tasks}
Four distinct classification tasks were designed using the SONAI dataset, based on the metadata accompanying each image or the anatomy shown.


\textbf{Plane Classification}\label{sec:plane_classification} assigns ultrasound images into four anatomical planes: abdomen (n=1947), brain (n=3059), femur (n=1832), and thorax (n=2125). This task is a core application in fetal ultrasound and serves as a primary in-distribution task for our experiments. \textbf{Scanner Classification} identifies the machine used to acquire each image, which can be GE Voluson S (n=3723), V830 (n=2282) or E10 (n=2958). \textbf{DICOM Type Classification} predicts image type, which can be a single-region b-mode (n=6786), multi-region b-mode (n=1964), or a color Doppler (n=213) ultrasound image. 
Finally, \textbf{Maternal BMI Group Classification} predicts the BMI group of the pregnant subject undergoing the ultrasound scan, i.e.~underweight (BMI $\leq$ 18, n=2452), normal (BMI 19-24, n=2260), overweight (BMI 25-29, n=2230), or obese (BMI $\geq$ 30, n=2021).

\section{Experiments and Results}
Following~\cite{mucsanyi2024benchmarking}, we trained 160 classifier models for the four classification tasks (see \autoref{sec:classification_tasks}),  using the eight UQ-methods (see \autoref{sec:uncertainty_quantification_methods}) repeated with five random seeds, i.e.~we test both i) the influence of the classification task, and ii) the UQ method independently. All models were based on a ResNet-50 backbone, and trained for 100 epochs. 
Experiments were conducted on an AlmaLinux 8.7 server with NVIDIA RTX A6000 GPU. All datasets follow a 80:10:10 train-val-test split. Statistical significance was tested using a multi-factor ANOVA model.


\begin{figure}[t]
    \centering
    \includegraphics[width=1\linewidth]{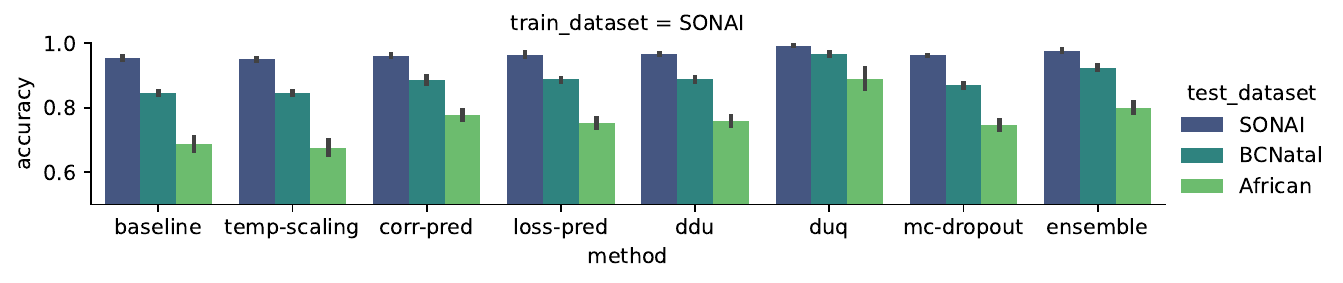}
    \caption{Accuracy of the anatomical plane classifiers, trained using the eight UQ-methods, in classifying anatomical planes across three test datasets.}
    \label{fig:hard_bma_accuracy_original}
\end{figure}

\subsection{Classification accuracy  drops as image characteristics change}
\label{sec:experiment_accuracy_sanity_check}
Before assessing uncertainty quantification, we validate the accuracy of our UQ-endowed classifiers when trained for our primary classification task: anatomical plane classification. As the models are trained on the SONAI dataset, we expect some distribution shift, and drop in accuracy from the SONAI to BCNatal test sets, as the latter comes from a different site. We expect a further drop for the African test set,  where we also expect a drop in image quality due to a less advanced scanner. 
Indeed, \autoref{fig:hard_bma_accuracy_original} shows a significant trend of accuracy dropping from the SONAI, to BCNatal, and to the African test sets for all models. 

This demonstrates the vulnerability of deep learning models to shifts in input image distribution, highlighting the importance of reliable OOD detection methods.  
This also motivates our subsequent experiments to evaluate the effectiveness of different OOD detection methods in identifying these input distribution shifts. 

\subsection{The effect of base classification task on OOD detection}

We now investigate our main research question: \textit{How does the base classification task affect the UQ-methods' performance in OOD detection?} We hypothesize that the task would influence the learned feature representations of the model and, consequently, their ability to identify OOD samples.

Moreover, we study our follow-up research question: \textit{Do "good" classification tasks perform equally well at detecting different plausible distribution shifts?}
To answer this, we study how the classification task affects our ability to detect two different families of distribution shifts that occur naturally in fetal ultrasound screening: Shift in image characteristics due to different (quality) scanners, and anatomical shift, which occurs when operators search for a given anatomical plane or feed an incorrect image to the model.

All models below were trained on the SONAI training set for each of the four classification tasks defined in \autoref{sec:classification_tasks}.

\begin{figure}[t]
    \centering
    \includegraphics[width=0.9\linewidth]{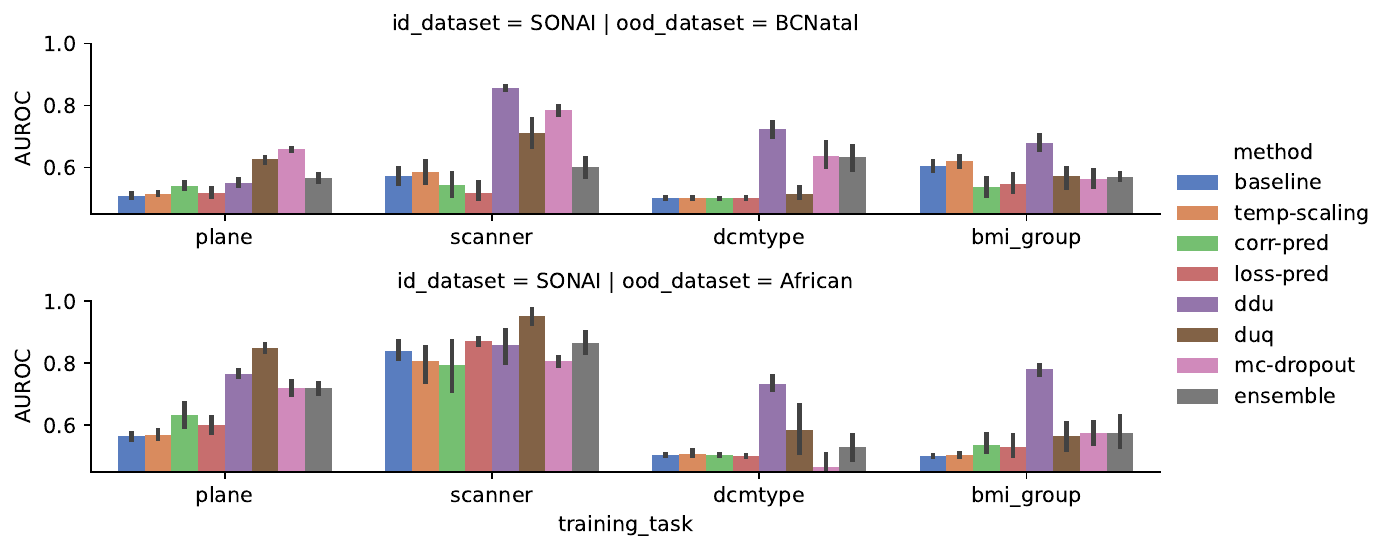}
    \caption{\textbf{Image characteristics shift:} OOD detection performance across UQ-methods, for the four classification tasks. The OOD datasets (BCNatal, African) constitute a shift in image characteristics (image quality, scanner characteristics).}
    \label{fig:varied_classification_task_oodness_image_quality_shift}
\end{figure}

\subsubsection{OOD 1: Shift in image characteristics.}
\label{sec:experiment_image_quality_shift}

\autoref{fig:varied_classification_task_oodness_image_quality_shift} shows how ID vs OOD classification performance, quantified via AUROC, distributes across the eight UQ-methods when built on the four different base classifiers. 

First, as expected, we see a significantly higher OOD classification performance for the African dataset than for BCNatal, confirming that most methods can pick up on the expected increased distribution shift for the African dataset. 

Second, we observe that for both distribution shifts, and for almost every UQ-method evaluated, models trained on the scanner classification task consistently and significantly outperformed models trained on the other three tasks. This suggests that, in the presence of a strong image quality shift, the scanner classification task led to feature representations that were remarkably more effective at distinguishing OOD data. 
Notably, models trained on the plane classification task, which represents our primary classification task of interest in the clinic, were almost consistently \emph{not} the best performing OOD detectors. This is particularly interesting, as the primary classifier would normally be the choice of base classifier for OOD detection.

\subsubsection{OOD 2: Anatomical shift.}
\label{sec:experiment_anatomical_feature_shift}
\begin{figure}[t]
    \centering
    \includegraphics[width=0.9\linewidth]{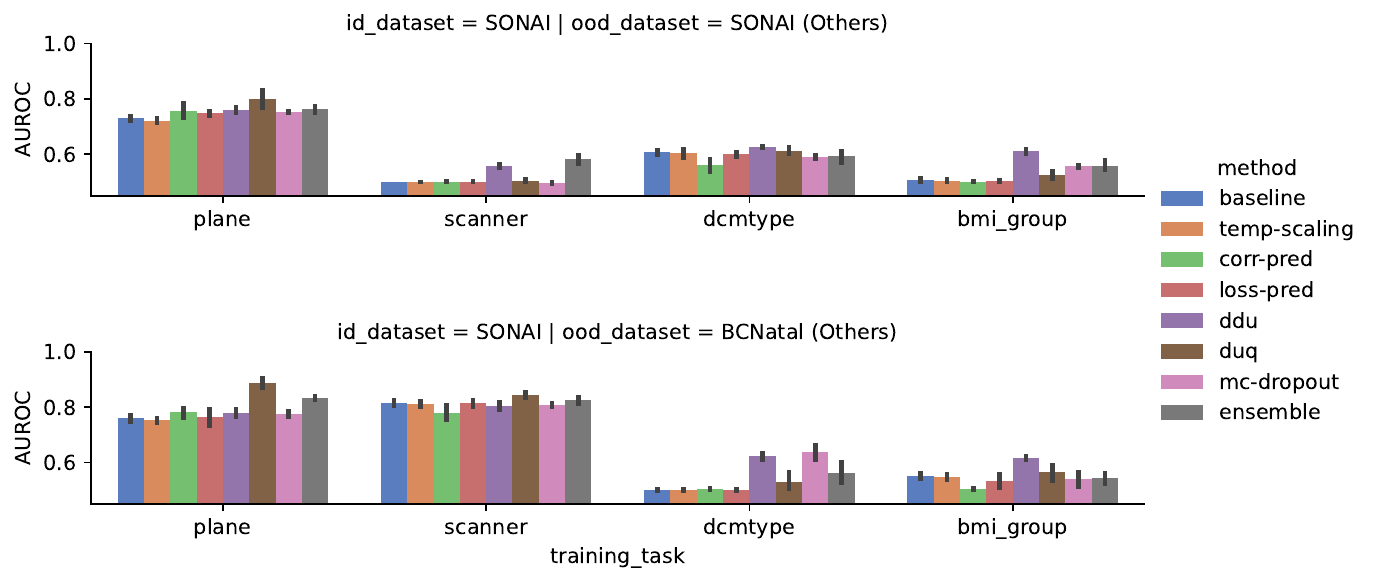}
    \caption{\textbf{Anatomical feature shift:} OOD detection performance across UQ-methods for each of the four classification tasks using the OOD datasets (SONAI (Others), BCNatal (Others)).}
    \label{fig:varied_classification_task_oodness_anatomical_feature_shift}
\end{figure}

\autoref{fig:varied_classification_task_oodness_anatomical_feature_shift} shows how ID vs OOD classification performance, quantified via AUROC, distributes across the eight UQ-methods built on the four base classifiers. We consider two different datasets with anatomical shift from the SONAI dataset; namely SONAI (Other) and BCNatal (Other).

This experiment yields a different picture from our first set of experiments: For detecting SONAI (Others), the UQ-methods built on the primary plane classification task perform best by far. This base classifier is also competitive on BCNatal (Other), although the scanner classifier is, again, slightly better.

\subsection{Abstained Prediction}
\label{sec:experiment_abstained_prediction}
\begin{figure}[t]
    \centering
    \includegraphics[width=0.9\linewidth]{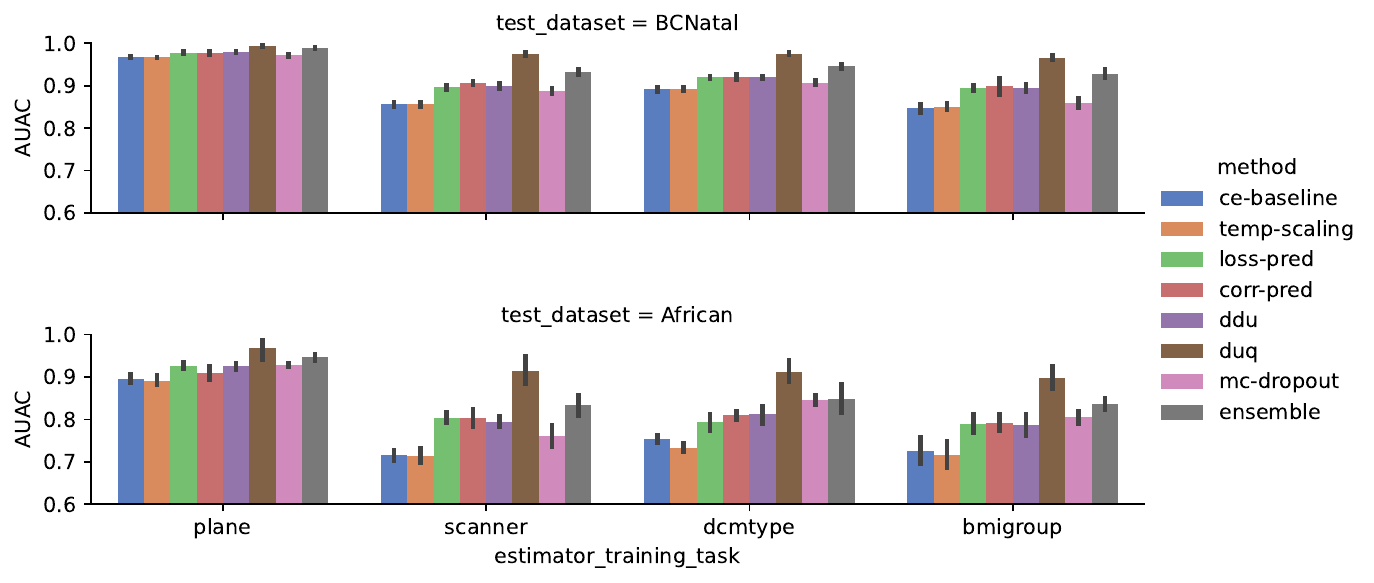}
    \caption{Performance of "plane classification" with abstained prediction, across base classifier for UQ-methods as well as image characteristic distribution shifts.}
    \label{fig:abstained_prediction}
\end{figure}

To further investigate the practical implications of task-dependent uncertainty quantification, we explored the performance of our primary (anatomical plane) classifier, using the different UQ-models in an abstained prediction scenario.  Abstained prediction is a more stringent evaluation of uncertainty estimation than OOD detection alone.  Here, the model abstains from making a prediction when $u(\mathbf{x}) > \tau$ for an input image $\mathbf{x}$ exceeds a predefined $\tau$, and accuracy is evaluated based on the remaining, non-abstained samples. 


We first utilized the models trained for the plane classification task with each of the eight UQ-methods. For each plane classification model, we varied the uncertainty threshold $\tau$, and calculated the accuracy of the model only on the samples for which it did not abstain (i.e., $u(\mathbf{x}) \leq \tau$).  We repeated this process for a range of thresholds $\tau$ to generate an accuracy coverage curve. Next, we repeated the experiment with $u(\mathbf{x})$ generated by the base models trained for scanner type, dicom type, and bmi group classification instead. Following the convention in \cite{mucsanyi2024benchmarking}, we evaluated the abstained prediction performance by calculating the area under the accuracy coverage curve (AUAC).

We had expected that models exhibiting strong OOD detection performance would also excel in abstained prediction. However, \autoref{fig:abstained_prediction} shows that models trained for scanner classification, which demonstrated superior OOD detection capabilities (especially for shifts in image characteristics), performed worse in abstained prediction for the plane classification task than UQ-models using plane classification as a base model: On both image characteristic shifts, uncertainty estimated with plane classification models generally leads to higher AUAC.

\section{Discussion and conclusion}

Our first result is that \textbf{the performance in OOD of a UQ-model depends heavily both the UQ-method's base classification task, and the OOD distribution shift.}
\autoref{sec:experiment_image_quality_shift} suggests that the base classification task plays a critical role in shaping the feature space utilized by the UQ methods. Different tasks lead to distinct learned feature spaces, ultimately impacting OOD detection. For image characteristic shifts (BCNatal and African datasets as OOD), scanner classification models excelled. This task trains models to be sensitive to texture and noise features directly affected by image quality degradation, making them ideal for detecting this type of OOD. For most anatomical shifts (SONAI (Other) and BCNatal (Other) datasets as OOD), on the other hand, plane classification models performed best. Training for anatomical plane recognition makes these models sensitive to broad anatomical differences, which is needed to identify images from entirely different anatomies as OOD. These further reinforce our central finding: OOD detection performance is not solely a function of the chosen uncertainty method, but is also determined by the interplay between the classification task and the specific ID-OOD shift.

Interestingly, scanner classification models also performed well in detecting anatomical shifts with BCNatal (Other) dataset as OOD. We hypothesize that this can be explained by considering the acquisition context of the BCNatal (Other) dataset. Some of the images in the BCNatal (Other) dataset were acquired with different ultrasound probes or acquisition protocols. This leads to detectable differences in image characteristics related to acquisition settings, which the scanner classification models are inherently sensitive to.

Our second important result is that \textbf{the best OOD performance does not imply the best abstinence performance.}
Our investigation into abstained prediction performance in \autoref{sec:experiment_abstained_prediction} revealed that optimal OOD detection performance does not necessarily translate to optimal performance for avoiding erroneous predictions in our clinical task of interest. Scanner classification models, which consistently demonstrated superior OOD detection across different ID-OOD shifts, actually underperformed in abstained prediction for our primary plane classification task compared to models trained directly for plane classification. This could reflect that a different datasets do not necessarily imply classifier failure. 
Models trained directly for plane classification, while potentially less sensitive to broader OOD shifts, appear to develop a more refined understanding of uncertainty within the plane classification task, enabling them to abstain more effectively while maintaining high accuracy on confident samples.

\textbf{Limitations.} First, our choice of UQ methods is limited. We have chosen to focus on UQ-methods that are easily available and widely used, while also being classifier-based to enable consistent experiments. Nevertheless, we hypothesize that our main point stands: There is no unversal UQ method to rule them all. 

Second, While OOD detection is usually approached using epistemic UQ,
we include methods that also measure aleatoric uncertainty. We argue, however, that this makes sense for our applications: While anatomical shifts and image characteristic shifts like artifacts and shadows are clearly epistemic, other image characteristic shifts, like blur or resolution, could be considered aleatoric. We thus find it appropriate also to include UQ methods with an aleatoric component.

\textbf{Conclusion.} Our findings have significant implications for the practical application and validation of uncertainty quantification. The choice of the most effective UQ-model is not universal and must be guided by the specific downstream application. For data shift monitoring systems, where the goal is to detect when the input data distribution has changed significantly enough to trigger model retraining, OOD detection performance should be prioritized. 
For clinical referral systems, where the aim is to automate the processing of confident cases while referring uncertain cases to clinicians, abstinence performance becomes the more critical metric 
for minimizing both errors in automated cases and clinician workload in abstained cases.

\clearpage

\bibliographystyle{splncs04}
\bibliography{mybibliography}

\end{document}